%% file: main.tex
\documentclass{article}
\usepackage{spconf,amsmath,graphicx}
\usepackage{adjustbox}
\input{macros}

\title{Utilizing Excess Resources in Training Neural Networks}
\name{Amit Henig and Raja Giryes}
\address{Electrical Engineering, Tel Aviv University}
\begin{document}
%
\maketitle
%


\input{00_abstract}

\input{01_intro}

\input{02_method}

\input{03_relation_to_other_works}

\input{04_experiments}

\input{05_conclusion}

\vfill\pagebreak

\bibliographystyle{IEEEbib}
\bibliography{strings,refs}

\end{document}

%% file: macros.tex
\usepackage{amsfonts}
\usepackage{xcolor}

\def\OurMethodFullName {Kernel Filtering Linear Overparameterization}
\def\OurMethod {KFLO}

\def\FC {fully connected}
\def\Conv {convolutional}
\def\FL {filtering layer}

\newcommand\mdsum[2]{\underset{#1}{\stackrel{#2}{\sum}}}
\newcommand\timesD[3]{#1 \times #2 \times (#3)}
\newcommand{\R}{\mathbb{R}}

\newcommand{\X}{\mathbb{X}}

\newcommand{\W}{\mathbb{W}}

\newif\ifdraft
\drafttrue

\ifdraft
\newcommand{\rgc}[1]{{\color{red}[\textbf{DC:} #1]}}
\newcommand{\duc}[1]{{\color{orange}[\textbf{RH:} #1]}}

\newcommand\TODO[1]{\textcolor{orange}{\textbf{TODO:} #1}}
\newcommand\DNOTE[1]{\textcolor{teal}{\textbf{DEBUG NOTE:} #1}}

\else
\newcommand{\rgc}[1]{}
\newcommand{\duc}[1]{}

\newcommand\TODO[1]{}
\newcommand\DNOTE[1]{}

\fi

%% file: 00_abstract.tex
\begin{abstract}
In this work, we suggest \textbf{\OurMethodFullName{} (\OurMethod{})}, where a linear cascade of \FL{}s is used during training to improve network performance in test time. We implement this cascade in a kernel filtering fashion, which prevents the trained architecture from becoming unnecessarily deeper. This also allows using our approach with almost any network architecture and let combining the \FL{}s into a single layer in test time. Thus, our approach does not add computational complexity during inference. We demonstrate the advantage of \OurMethod{} on various network models and datasets in supervised learning.
\end{abstract}

%% file: 01_intro.tex
\section{Introduction}
\label{sec:intro}

Deep neural networks have shown remarkable capabilities in computer vision and natural language processing. Research on improving the deployed networks' performance has been done in many fields and different problem frameworks. In this work, we focus on the common case where the deployed model architecture is chosen in advance of training, and resources are abundant during training relative to inference time. These resources can be used to improve performance.
Specifically, we focus on using a larger network during training that generates a smaller one for inference. 

\begin{figure}[htb]
\begin{minipage}[b]{0.4\linewidth}
  \centering
  \centerline{\includegraphics[width=8cm]{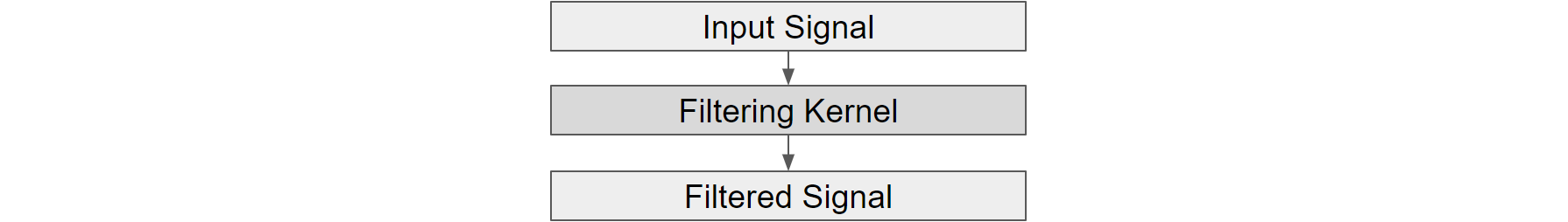}}
  \centerline{(a) Vanilla Filtering Layer}\medskip
\end{minipage}
\begin{minipage}[b]{0.45\linewidth}
  \centering
  \rightline{\includegraphics[width=8cm]{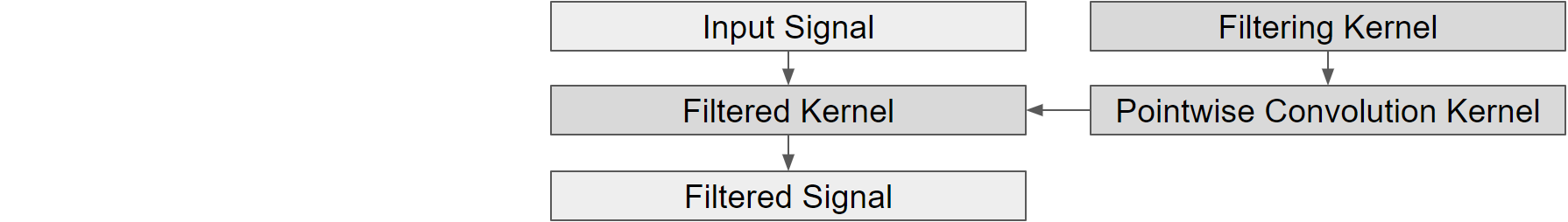}}
  \centerline{\hspace{-0.5in}(b) \OurMethod{} Filtering Layer}\medskip
\end{minipage}
\hfill
\begin{minipage}[b]{1.0\linewidth}
  \centering
  \centerline{\includegraphics[width=8.5cm]{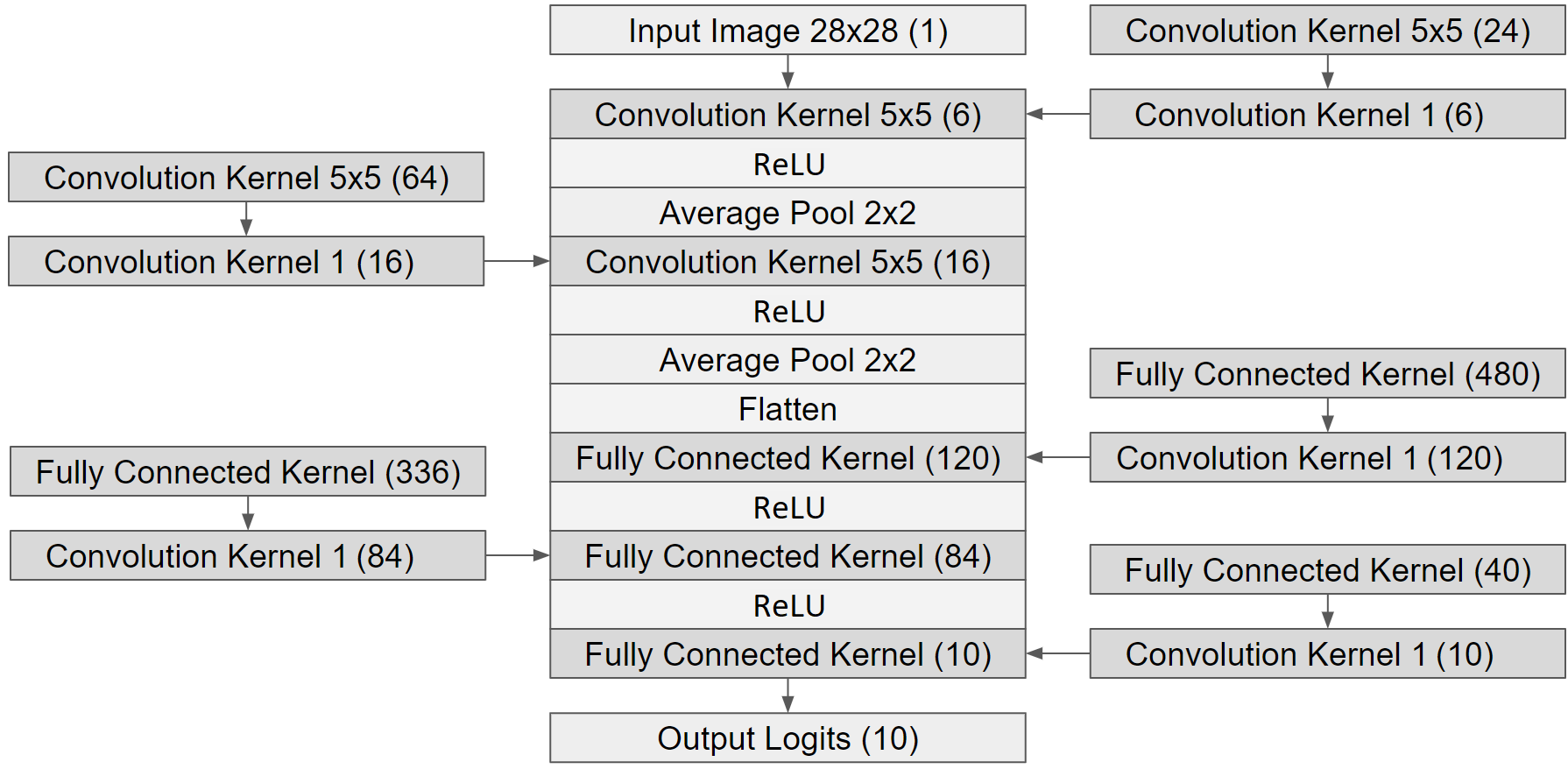}}
  \centerline{(c) LeNet-5 with \OurMethod{}}\medskip
\end{minipage}
\vspace{-0.3in}
\caption{A brief preview of our method. Layers containing filtering weights are marked with a darker shade. (a) A common vanilla filtering layer. (b) Filtering layer with \OurMethod{}: The kernel that filters the input signal is the result of a pointwise convolution applied on a different filtering kernel. The weights outside the input signal's path can be discarded after training, deploying only the Filtered Kernel. (c) LeNet-5 implemented with \OurMethod{}, for 28x28 grayscale images of 10 possible classes. Here too, the weights outside the signal's main path can be discarded after training, resulting in the original LeNet-5 architecture to be used at inference time.} \vspace{-0.2in}
\label{fig:BriefPreview}
\end{figure}

One may treat this problem setting as a network compression or a knowledge transfer problem.
A known strategy for this task is Knowledge Distillation (KD). There, knowledge is transferred from a pre-trained teacher network to a desired student network using signals from the teacher. Interestingly, the student can match or even outperform the teacher’s performance while being notably less complex.
KD was first introduced by \cite{ref:FirstKd} for model compression and generalized by Hinton et. al. \cite{ref:HintonKd}. Various KD extensions exist such as KD in generations \cite{ref:BornAgainNN}, ensemble of students teaching each other \cite{ref:DeepMutualLearning}, the teacher assistant \cite{ref:TeacherAssistant}, and many more.

A different approach suggests using structural reparameterization, term denoted by \cite{ref:RepVGG} for transforming one architecture's parameters to produce the parameters used by another architecture.
Specifically, in the context of this paper, a collapsible overparameterization. Common networks are typically overparameterized, yet numerous empirical results show that even deeper and/or wider networks can achieve better performance \cite{ref:WRN}. Recent works have suggested that this does not always result from stronger computational capabilities of the bigger network compared to the smaller one. The smaller one, which can have enough expressive power to memorize the dataset, is usually harder to train, that is, it is harder for the smaller network to achieve a good generalization as the bigger network; however, with the right training it might \cite{ref:TeacherAssistant, ref:RethinkGen}. Structural re-parameterization methods can improve performance by adding overparameterization to a desired architecture, while dropping the excess weights after training, so the deployed network at inference time is of the original desired architecture. 
We focus on linear overparameterization (LO) as most structural re-parameterization relies on linearity.

Linear networks and their convergence with gradient decent have been studied for a long time. It has been shown that using a deep linear neural network can have advantages over the use of a single linear layer, despite having the same computational power, e.g., by accelerating convergence  \cite{ref:DeepLinearOver}. 
This work also notes the vanishing gradient problem that can occur in deep linear networks with near zero initialization.

A logical next step is to introduce LO to regular non linear networks by adding redundant (transformable / collapsible) linear layers.
ExpandNets \cite{ref:ExpandNets} used LO by expanding every 2D convolution to multiple consecutive convolutions, keeping the same receptive field size.
While being promising on shallow networks, it is less effective and even harmful for deep networks. Also, even for shallow networks, too deep LO can harm performance due to vanishing gradients \cite{ref:ExpandNets, ref:DeepLinearOver}.

\noindent {\bf Contribution.} In this paper, we propose a novel framework, \emph{\OurMethodFullName{} (\OurMethod{})}, for improving the performance of a target network.
Our approach attempts to better utilize linear overparameterization and handle the vanishing gradient problem, while also being compatible with every type of filtering layer and staying easy to implement.
Conceptually redundant linear layers are added to the network in a kernel filtering fashion, which does not make the network deeper from the network's input perspective. The overparameterized weights can be discarded after being used in training, thus making inference time light-weight.
Our method enjoys a much smaller number of calculations needed during training compared to using a straightforward deeper LO.
A a brief preview of \OurMethod{} can be found in Figure \ref{fig:BriefPreview}.
Code is avavilable at https://github.com/AmitHenig/KFLO.

%% file: 02_method.tex
\section{Method}
\label{sec:method}

{\bf Notations.}
An input/signal/feature-map tensor has the shape $[batch \times channels \times spatial]$ (the $spatial$ dimensions of an image for example will be $Height \times Width$). The kernel of a \FL{} (such as convolution or \FC{}) has the shape $[output\_channels \times input\_channels \times spatial']$. The kernel is regarded to have $output\_channels$ filters, each of shape $[1 \times input\_channels \times spatial']$.
From this point forward, we will examine a 2D \Conv{} layer as our \FL{}, though our solution can easily be extended to any spatial dimension (as we will see later), and \FC{} layers can be treated as inputs and filters having spatial dimensions sizes of 1. Furthermore, for simplicity, we will omit the bias, as it can be easily added.

\noindent {\bf LO.}
Consider a linear cascade comprised of two 2D \Conv{} layers  with corresponding kernels/weights $\{\W_i\}_{i=1}^2$, where the second layer is a pointwise convolution with a trivial sliding window.
This cascade is a LO of some 2D \Conv{} layer with weights $\W'$ and the same sliding window properties as $\W_1$.
Meaning, that for the same input, kernel $\W'$ gives the same output as the cascade $\{\W_i\}_{i=1}^2$:
$$\W' * \X_1 = \W_2 * (\W_1 * \X_1) = \W_2 * \X_2 = \X_3$$
\begin{align*}
&\X_1 \in \R^{\timesD{batch}{ch_1}{H \times W}}&&;&& \W_1 \in \R^{\timesD{ch_2}{ch_1}{M \times N}} \\
&\X_2 \in \R^{\timesD{batch}{ch_2}{H' \times W'}}&&;&& \W_2 \in \R^{\timesD{ch_3}{ch_2}{1 \times 1}} \\
&\X_3 \in \R^{\timesD{batch}{ch_3}{H' \times W'}}&&;&& \W' \in \R^{\timesD{ch_3}{ch_1}{M \times N}}
\end{align*}
Note that neither $H'$, $W'$ nor the receptive field size are affected by the pointwise convolution, and so are solely dependent on the properties of $\W_1$.
This LO equality holds when:
\begin{eqnarray}
\label{eq:t_wt2}
\W'(\delta_3, \delta_1, m, n) = \mdsum{\delta_2}{ch_2} \W_2(\delta_3, \delta_2, 0, 0) \cdot \W_1(\delta_2, \delta_1, m, n)
\end{eqnarray}
This can be shown by rearranging the convolution:
\begin{align*}
\X_3(b, \delta_3, h', w') = \mdsum{\delta_2}{ch_2} \W_2(\delta_3, \delta_2, 0, 0) \cdot \X_2(b, \delta_2, h', w')
\end{align*}
where $\X_2(b, \delta_2, h', w')$ is
\begin{align*}
\mdsum{\delta_1}{ch_1} \mdsum{m}{M} \mdsum{n}{N} \W_1(\delta_2, \delta_1, m, n) \cdot \X_1(b, \delta_1, \mu^{m, n}_{(h', w')}, \nu^{m, n}_{(h', w')})
\end{align*}
and $\mu$, $\nu$ are the kernel-to-input spatial mapping functions, defined by the sliding window properties (e.g., the mapping functions for a trivial sliding window are $\mu=h'+m$ and $\nu=w'+n$).
Eq. \ref{eq:t_wt2} shows that each filter in $\W'$ is a combination of $\W_1$ filters, weighted by a corresponding filter in $\W_2$.

\noindent {\bf Kernel Filtering.}
We introduce our kernel filtering by transforming Eq. \ref{eq:t_wt2} to a \Conv{} representation. 
Given the nature of the pointwise convolution, kernel filtering for this case may be represented with a 1D convolution; making its implementation independent of the number of spatial dimensions.
This is done by convolution with the reshaped kernels. For some spatial location $t=(M \cdot N) \cdot \delta_1 + N \cdot m + n$:
\begin{alignat*}{2}
&\W_1^R \in \R^{1 \times ch_2 \times (ch_1 \cdot M \cdot N)} &;&\quad \W_1^R(0, \delta_2, t) = \W_1(\delta_2, \delta_1, m, n)
\\ &\W_2^R \in \R^{ch_3 \times ch_{2} \times (1)} &;&\quad \W_2^R(\delta_3, \delta_2, 0) = \W_2(\delta_3, \delta_2, 0, 0)
\\ &\W'^R \in \R^{1 \times ch_3 \times (ch_1 \cdot M \cdot N)} &;&\quad \W'^R(0, \delta_3, t) = \W'(\delta_3, \delta_1, m, n)
\end{alignat*}
$$\W'^R(0, \delta_3, t) = \mdsum{\delta_2}{ch_2} \W_2^R(\delta_3, \delta_2, 0) \cdot \W_1^R(0, \delta_2, t)$$
We get a 1D pointwise convolution $\W_2^R$ for input $\W_1^R$ with batch size of 1, resulting in the output $\W'^R$.
Getting $\W'$ from $\W'^R$ is of course done by reshaping.
$$\W' * \X_1 = (\W'^R)^R * \X_1 = (\W_2^R * \W_1^R)^R * \X_1 = \X_3$$
\noindent {\bf \OurMethod{}.}
It can be shown that the above LO and kernel filtering equivalences hold for any number of pointwise kernels used in the linear cascade $\{\W_i \in \R^{\timesD{ch_{i+1}}{ch_{i}}{1 \times 1}}\}_{i=2}^B$, reshaping them in the same manner to 1D pointwise convolutions:
\begin{alignat}{2}
\label{eq:t_y1}
&\X_{B+1} &&= \W_B * (\W_{B-1} * ... * (\W_2 * (\W_1 * \X_1)) ... ) \\
\label{eq:t_y2}
& &&=  (\W_B^R * (\W_{B-1}^R * ... * (\W_2^R * \W_1^R) ... ))^R * \X_1 \\
\label{eq:t_y3}
& &&= (\W'^R)^R * \X_1 = \W' * \X_1
\end{alignat}
For a deployed kernel $\W' \in \R^{\timesD{ch_{out}}{ch_{in}}{M \times N}}$, our method is implemented during training by creating weights $\{\W_i\}_{i=1}^B$ and calculating the effective convolution used in the network using Eqs. \ref{eq:t_y2}, \ref{eq:t_y3}.
After training is complete, the LO weights $\{\W_i\}_{i=1}^B$ are discarded, only keeping $\W'$ to be used at inference time (see Figure \ref{fig:BriefPreview}).
Applying \OurMethod{} on a depthwise convolution layer is done exactly as for conventional convolution, keeping in mind that the number of $\W_1$ input channels is $\frac{\#input\_channels}{\#groups}$.
We define a scalar \textbf{width multiplier} $\pmb{\rho}$, and set all configurable LO widths $\{ch_i\}_{i=2}^B$ to $Round(\rho \cdot ch_{out})$. We also denote $\pmb{B}$ as the \textbf{linear depth multiplier}.
\OurMethod{} for a desired deployed architecture is implemented by defining $B$$>$$1$ and $\rho$$>$$0$ (mainly choosing values greater than 1 for LO), and applying our method as described above for every \FL{} in the network (example with $B$=2, $\rho$=4 in Figure \ref{fig:BriefPreview}.c).


We present two explanations for KFLO success. First, we know that overparameterization helps train networks in general. Second, from fully linear deep networks we know that this redundant overparameterization can improve convergence.
Comparing to a LO with vanilla feature filtering, \OurMethod{} does not make the trained architecture deeper, which is a big advantage as a common problem of deep LO/networks is the vanishing gradient which can really harm performance. Also, our kernel filtering gives a great reduction in the number of operations that LO adds. While in feature filtering each LO layer is applied to the full spatial size of the layer's input tensor, with our kernel filtering the LO layers filter a tensor with spatial size of a convolution layer, which is usually significantly smaller than that of the layer's input signal.

%% file: 03_relation_to_other_works.tex
\section{Relation to other LO works}
DO-Conv \cite{ref:DOConv} has a similar approach as ours. They utilize a linear depthwise separable \Conv{} layer, showing that it can be collapsed to a single convolution kernel. Similarly to us, DO-Conv suggests a kind of kernel filtering (kernel composition), to replace the vanilla feature filtering during training. 
There are two key differences between our method and DO-Conv. First, the linear \Conv{} cascade of our method is different than the linear depthwise separable convolution in DO-Conv. 
Second, while we performs kernel filtering by passing forward $\W_1^R$ through the 1D linear cascade, DO-Conv passes the pointwise kernel backwards by filtering it with a transposed depthwise convolution.
The work in Orthogonal Over-Parameterized Training \cite{ref:OrthoOPT} uses a tailored LO filtering weights, with further restrictions to enforce an orthogonal transformation on static weights.
ACNet \cite{ref:ACNet} suggests Asymmetric Convolution Block (ACB), which can be applied on a square convolution kernel. During training the kernel is replaced with three parallel paths, each with a convolution kernel followed by batch normalization, and their outputs are summed. In two of the paths the kernels are of 1D, one horizontal and the other vertical.
This method does not make the network deeper during training in terms of filtering kernels and they show improvement in both small and big networks.
Note that the commonly used pair of convolution followed by batch normalization can also act as a LO on its own. 
Diverse Branch Block (DBB) \cite{ref:DBB} suggests LO by replacing the convolution kernel with an Inception-like DBB instance during training. ACB (ACNet) can be viewed as a special case of DBB.
LO in training was also utilized in \cite{ref:RepVGG,ref:CollapsibleLB,ref:WeightedAvgKernels}.
The work in \cite{ref:FewSKD} uses LO during training alongside block-level KD on a student network initially obtained by compressing the pre-trained teacher. LO is applied in a vanilla feature filtering fashion by adding a pointwise convolution solely at the end of each defined block and is collapsed/merged after training. Compared to them, our kernel filtering is applied on every \FL{} with further wider LO (width multiplier $\rho$$>$$1$).

%% file: 04_experiments.tex
\section{Experiments}
\label{sec:experiments}

In this section we demonstrate our approach on two popular datasets: CIFAR-10 and CIFAR-100 \cite{ref:Cifar}. We evaluate \OurMethod{} and compare it to both the vanilla approach and other relevant training methods. We present experimental results for different architectures and datasets, showing the advantages of our method. We also compare the performance of \OurMethod{} with different linear depth multipliers, $B$, and different width multipliers, $\rho$. Finally, we show that \OurMethod{} can have a similar effect of transfer learning.
Note that with infinite memory and time, many methods can be combined and applied simultaneously, multiple times. 
However, in the scope of this paper we will only refer to training with no more than one LO method applied at a time.


\noindent \textbf{Networks.}
WRN-D-K will denote a Wide Residual Network \cite{ref:WRN}, with the wide-dropout residual block, of depth $D$ 
(with a $3$ residual groups structure) 
and widening factor $K$.
In addition, we test with Cifar-quick \cite{ref:CifarQuick} and VGG16 \cite{ref:VGG16} where the two \FC{} layers are replaced with global average pooling followed by a \FC{} layer that outputs 512 channels, as \cite{ref:ACNet}. For both architectures, batch normalization were added after every convolution layer, again as \cite{ref:ACNet}.

\noindent \textbf{Implementation Details.}
When using our training method, \OurMethod{} is applied to all the \Conv{} and \FC{} kernels.
\OurMethod{} with a linear depth multiplier $B$ and width multiplier $\rho$ will be denoted by \textbf{\OurMethod{} Bx}$\pmb{\rho}$.
We apply a weak weight decay of $10^{-9}$ on the pointwise convolutions in the linear cascades $\{\W_i\}_{i=2}^B$, and initialize them with identity (dirac).
We apply the vanilla training weight decay value on the generated weights $\W'$, and no weight decay on the filtered kernels $\W_{1}$, thus not adding a hyper parameter to tune.
DO-Conv is implemented with $D_{mul}$ equals to the filter's spatial size $M \times N$, as in \cite{ref:DOConv}.
Exponential moving average of the network weights when trained in a vanilla fashion will be referred to as \textbf{EMA}.
The Experiments ran on NVIDIA GeForce RTX 2080 Ti. We followed the experimental settings of \cite{ref:ACNet}, excluding WRN Experiments where we followed \cite{ref:DeepMutualLearning}.

\noindent \textbf{Transfer learning (TL)} experiments were conducted on CIFAR-100, with only 40\% of the training data available (uniformly sampled from each class). Networks pre-trained in the vanilla fashion on the full CIFAR-10 training set were used for weights initialization. In the classic strategy, a pre-trained network is used as the starting point and it is fine-tuned with a learning rate of an order of magnitude smaller.

With \OurMethod{}, more than one pre-trained model can be utilized in weight initialization. Setting the width multiplier $\rho$ to the number of pre-trained networks results in \OurMethod{} filtering kernels ($\W_{1}$) wide enough to hold all relevant pre-trained kernel filters.
Before Training starts, we stack the corresponding pre-trained kernels and use them as initialization weights for the relevant filtering kernels. Layers on which \OurMethod{} is not applied are initialized with the weights of the first pre-trained network. This with the Identity initialization of \OurMethod{}'s LO pointwise convolutions, means that the starting network gives identical outputs to the first pre-trained network.

\subsection{Results}
\label{ssec:exp_res}

\noindent {\bf Comparison.} We first present results comparing our method to vanilla training, EMA, ACNet \cite{ref:ACNet} and DO-Conv \cite{ref:DOConv}. Results for CIFAR-10 and CIFAR-100 can be found in Tables \ref{table:resultsCifar10Compare} and  \ref{table:resultsCifar100Compare}, respectively.
We can see that \OurMethod{} achieves the highest accuracy with all featured architectures.
\input{tables/results_cifar10_compare}
\input{tables/results_cifar100_compare}

\noindent {\bf Ablation study.} We compare the performance of our method with different linear depth multipliers $B$, and different width multipliers $\rho$. We show results for CIFAR-10 in Table~\ref{table:resultsCifar10AbStudy} and for CIFAR-100 in Table~\ref{table:resultsCifar100AbStudy}. It seems that in most cases, the bigger the width multiplier the better the performance boost is. However, a larger depth multiplier harms performance.
\input{tables/results_cifar10_abStudy}

\noindent {\bf Transfer learning.} Table~\ref{table:resultsTransferLearning} shows results for the TL problem. It displays results for both networks trained from scratch, and networks trained with TL - denoted with the TL suffix. We can see the expected performance boost TL gives vanilla training. 
Yet, it would appear that \OurMethod{} does not benefit much from the pre-trained weights, and sometimes it can even harm performance. It seems that the straight forward method to incorporate the pre-trained networks with \OurMethod{} is not optimal. Nonetheless, \OurMethod{} (without TL) outperforms vanilla TL, and it would seem that it achieves a similar effect although it does not have access to the extra data (CIFAR-10).

%% file: tables/results_cifar10_compare.tex
\begin{table}
\begin{adjustbox}{width=1.\linewidth}
\begin{tabular}{|l|l|l|l|l|l|l|}
\hline
Method      & Cifar-quick         & VGG16          & {\small wResnet-16-8 }        & {\small wResnet-28-2 }        &  {\small wResnet-28-5 }        & {\small wResnet-28-10}        \\ \hline
vanilla     & 86.26 ±0.22         & 93.85 ±0.18    & 95.44 ±0.12          & 94.49 ±0.12          & 95.50 ±0.06          & 95.85 ±0.10          \\
EMA         & 86.26 ±0.21         & 93.87 ±0.17    & 95.44 ±0.12          & 94.52 ±0.10          & 95.49 ±0.05          & 95.84 ±0.09          \\
ACNet       & 86.83 ±0.28         & 94.41 ±0.12    & 95.62 ±0.14          & 94.75 ±0.18          & 95.79 ±0.14          & 96.14 ±0.06          \\
DO-Conv & 86.85 ±0.14         & 94.00 ±0.21    & 95.50 ±0.14          & 94.72 ±0.09          & 95.61 ±0.15          & 96.01 ±0.12          \\
KFLO 2x4    & \textbf{87.41 ±0.23} & \textbf{94.70 ±0.09} & \textbf{95.68 ±0.07} & \textbf{94.89 ±0.23} & \textbf{95.89 ±0.09} & \textbf{96.22 ±0.12} \\ \hline
\end{tabular}
\end{adjustbox}
\caption{CIFAR-10 accuracy results, methods comparison.}
\label{table:resultsCifar10Compare}
\end{table}

%% file: tables/results_cifar100_compare.tex
\begin{table}
\begin{adjustbox}{width=1.\linewidth}
\begin{tabular}{|l|l|l|l|l|l|l|}
\hline
Method      & Cifar-quick         & VGG16          & {\small wResnet-16-8 }        & {\small wResnet-28-2 }        & {\small wResnet-28-5}         & {\small wResnet-28-10}        \\ \hline
vanilla     & 57.80 ±0.30         & 73.97 ±0.28    & 78.58 ±0.22          & 75.37 ±0.26          & 78.89 ±0.39          & 80.18 ±0.25          \\
EMA         & 57.80 ±0.26         & 73.97 ±0.29    & 78.60 ±0.21          & 75.44 ±0.27          & 78.88 ±0.39          & 80.20 ±0.25          \\
ACNet       & 58.38 ±0.39         & 74.95 ±0.22    & 79.10 ±0.18          & 76.02 ±0.32          & 79.64 ±0.32          & 81.36 ±0.14          \\
DO-Conv & 58.86 ±0.20         & 74.38 ±0.17    & 79.01 ±0.28          & 75.73 ±0.29          & 79.10 ±0.19          & 80.59 ±0.29          \\
KFLO 2x4    & \textbf{59.78 ±0.25} & \textbf{75.69 ±0.14} & \textbf{79.74 ±0.24} & \textbf{76.07 ±0.29} & \textbf{79.96 ±0.27} & \textbf{81.38 ±0.15} \\ \hline
\end{tabular}
\end{adjustbox}
\caption{CIFAR-100 accuracy results, methods comparison.}
\label{table:resultsCifar100Compare}
\end{table}

%% file: tables/results_cifar10_abStudy.tex
\begin{table}
\begin{adjustbox}{width=1.\linewidth}
\begin{tabular}{|l|l|l|l|l|}
\hline
Method   & {\small wResnet-16-8}         & {\small wResnet-28-2}         & {\small wResnet-28-5}         & {\small wResnet-28-10}        \\ \hline
KFLO 2x1 & 95.50 ±0.17          & 94.59 ±0.10          & 95.83 ±0.09          & 96.16 ±0.13          \\
KFLO 2x2 & 95.62 ±0.11          & 94.60 ±0.18          & 95.84 ±0.13          & \textbf{96.28 ±0.10} \\
KFLO 2x3 & 95.59 ±0.10          & 94.88 ±0.08          & 95.79 ±0.15          & 96.26 ±0.15          \\
KFLO 2x4 & \textbf{95.68 ±0.07} & \textbf{94.89 ±0.23} & \textbf{95.89 ±0.09} & 96.22 ±0.12          \\
KFLO 3x2 & 94.65 ±0.15          & 91.98 ±0.14          & 94.68 ±0.17          & 95.65 ±0.11          \\ \hline
\end{tabular}
\end{adjustbox}
\caption{CIFAR-10 accuracy results, \OurMethod{} ablation study.}
\label{table:resultsCifar10AbStudy}
\vspace{1em}
\begin{adjustbox}{width=1.\linewidth}
\begin{tabular}{|l|l|l|l|l|}
\hline
Method   & {\small wResnet-16-8}         & {\small wResnet-28-2}         &  {\small wResnet-28-5}         & {\small wResnet-28-10}        \\ \hline
KFLO 2x1 & 79.47 ±0.21          & 75.69 ±0.39          & 79.60 ±0.24          & 81.35 ±0.23          \\
KFLO 2x2 & 79.68 ±0.31          & 75.91 ±0.31          & 79.76 ±0.22          & 81.36 ±0.15          \\
KFLO 2x3 & 79.59 ±0.16          & 76.00 ±0.29          & \textbf{79.96 ±0.18} & \textbf{81.46 ±0.15} \\
KFLO 2x4 & \textbf{79.74 ±0.24} & \textbf{76.07 ±0.29} & \textbf{79.96 ±0.27} & 81.33 ±0.20          \\
KFLO 3x2 & 76.51 ±0.26          & 68.21 ±0.39          & 75.69 ±0.22          & 78.93 ±0.12          \\ \hline
\end{tabular}
\end{adjustbox}
\caption{CIFAR-100 accuracy results, \OurMethod{} ablation study.}
\label{table:resultsCifar100AbStudy}
\vspace{1em}
\begin{adjustbox}{width=1.\linewidth}
\begin{tabular}{|l|l|l|l|l|}
\hline
Method                     & {\small wResnet-16-8}         & {\small wResnet-28-2}         & {\small wResnet-28-5}         &  {\small wResnet-28-10}        \\ \hline
vanilla                    & 69.81 ±0.19          & 66.69 ±0.24          & 69.56 ±0.19          & 70.79 ±0.41          \\
vanilla TL  & 71.20 ±0.41          & 67.12 ±0.25          & 71.43 ±0.42          & 73.15 ±0.33          \\
KFLO 2x1                   & 71.50 ±0.23          & 67.06 ±0.53          & 71.04 ±0.27          & 73.11 ±0.26          \\
KFLO 2x1 TL & 71.62 ±0.37          & 67.26 ±0.41          & 71.33 ±0.34          & 72.19 ±0.35          \\
KFLO 2x4                   & 71.86 ±0.14          & 67.66 ±0.34          & \textbf{71.63 ±0.27} & \textbf{73.36 ±0.29} \\
KFLO 2x4 TL & \textbf{71.97 ±0.32} & \textbf{67.85 ±0.32} & 71.60 ±0.22          & 72.55 ±0.26          \\ \hline
\end{tabular}
\end{adjustbox}
\caption{Transfer Learning accuracy results. Only 40\% of the CIFAR-100 training set available. The TL suffix denotes using networks pre-trained on the full CIFAR-10 training set.}
\label{table:resultsTransferLearning}
\end{table}

%% file: 05_conclusion.tex
\section{Conclusion}
\label{sec:conclusion}

This work introduced \OurMethod{}, a novel approach to train a neural network 
in a structural re-parameterization fashion 
by modeling kernels as the result of linear \Conv{} kernel filtering. Interestingly, this over-parameterization, which can create a very large redundancy, improves the network training. This stands in line with the recent both practical and theoretical findings that overparameterization improves the generalization of neural networks 
\cite{Allen19Learning,neyshabur2018the,Venturi19Spurious,Brutzkus2019WhyDL}.

We have focused on demonstrating our approach in supervised learning and transfer learning settings, though we believe that our proposed framework has large potential in being applied to other problems such as domain adaptation, semi-supervised learning and incremental learning.  
Moreover, one may consider in our scheme many other topologies for the kernel filtering block. We hope that this new tool with its different variations will be used to improve many other tasks. Its simplicity allows using and extending it to new setups easily.

\noindent {\bf Acknowledgement.} This research was supported by Wipro and ERC-StG grant no. 757497

%% file: main.bbl
\begin{thebibliography}{10}

\bibitem{ref:FirstKd}
Cristian Bucila, Rich Caruana, and Alexandru Niculescu-Mizil,
\newblock ``Model compression,''
\newblock in {\em KDD '06}, 2006.

\bibitem{ref:HintonKd}
Geoffrey Hinton, Oriol Vinyals, and Jeff Dean,
\newblock ``Distilling the knowledge in a neural network,''
\newblock in {\em NIPS Deep Learning Workshop}, 2014.

\bibitem{ref:BornAgainNN}
Tommaso Furlanello, Zachary~Chase Lipton, Michael Tschannen, Laurent Itti, and
  Anima Anandkumar,
\newblock ``Born again neural networks,''
\newblock in {\em ICML}, 2018.

\bibitem{ref:DeepMutualLearning}
Y.~{Zhang}, T.~{Xiang}, T.~M. {Hospedales}, and H.~{Lu},
\newblock ``Deep mutual learning,''
\newblock in {\em Conference on Computer Vision and Pattern Recognition
  (CVPR)}, 2018, pp. 4320--4328.

\bibitem{ref:TeacherAssistant}
Seyed-Iman Mirzadeh, Mehrdad Farajtabar, Ang Li, Nir Levine, Akihiro Matsukawa,
  and Hassan Ghasemzadeh,
\newblock ``Improved knowledge distillation via teacher assistant,''
\newblock in {\em AAAI}, 2020, pp. 5191--5198.

\bibitem{ref:RepVGG}
Xiaohan Ding, X.~Zhang, Ningning Ma, Jungong Han, Guiguang Ding, and Jian Sun,
\newblock ``Repvgg: Making vgg-style convnets great again,''
\newblock in {\em CVPR}, 2021.

\bibitem{ref:WRN}
Sergey Zagoruyko and Nikos Komodakis,
\newblock ``Wide residual networks.,''
\newblock in {\em British Machine Vision Conference}, 2016.

\bibitem{ref:RethinkGen}
C.~Zhang, S.~Bengio, Moritz Hardt, B.~Recht, and Oriol Vinyals,
\newblock ``Understanding deep learning requires rethinking generalization.,''
\newblock in {\em ICLR}, 2017.

\bibitem{ref:DeepLinearOver}
Sanjeev Arora, N.~Cohen, and Elad Hazan,
\newblock ``On the optimization of deep networks: Implicit acceleration by
  overparameterization,''
\newblock {\em ICML 2018}, 2018.

\bibitem{ref:ExpandNets}
Shuxuan Guo, Jose~M. Alvarez, and M.~Salzmann,
\newblock ``Expandnets: Linear over-parameterization to train compact
  convolutional networks,''
\newblock {\em NeurIPS}, 2020.

\bibitem{ref:DOConv}
Jinming Cao, Yangyan Li, M.~Sun, Ying Chen, D.~Lischinski, D.~Cohen-Or,
  B.~Chen, and C.~Tu,
\newblock ``Do-conv: Depthwise over-parameterized convolutional layer,''
\newblock {\em ArXiv}, vol. abs/2006.12030, 2020.

\bibitem{ref:OrthoOPT}
Weiyang Liu, Rongmei Lin, Z.~Liu, J.~Rehg, Li~Xiong, and L.~Song,
\newblock ``Orthogonal over-parameterized training,''
\newblock {\em ArXiv}, vol. abs/2004.04690, 2020.

\bibitem{ref:ACNet}
Xiaohan Ding, Yuchen Guo, Guiguang Ding, and J.~Han,
\newblock ``Acnet: Strengthening the kernel skeletons for powerful cnn via
  asymmetric convolution blocks,''
\newblock {\em 2019 IEEE/CVF International Conference on Computer Vision
  (ICCV)}, pp. 1911--1920, 2019.

\bibitem{ref:DBB}
Xiaohan Ding, Xiangyu Zhang, Jungong Han, and Guiguang Ding,
\newblock ``Diverse branch block: Building a convolution as an inception-like
  unit,''
\newblock {\em ArXiv}, vol. abs/2103.13425, 2021.

\bibitem{ref:CollapsibleLB}
K.~Bhardwaj, M.~Milosavljevic, A.~Chalfin, Naveen Suda, Liam O'Neil, Dibakar
  Gope, Lingchuan Meng, Ramon~Matas Navarro, and Danny Loh,
\newblock ``Collapsible linear blocks for super-efficient super resolution,''
\newblock {\em ArXiv}, vol. abs/2103.09404, 2021.

\bibitem{ref:WeightedAvgKernels}
Shoufa Chen, Y.~Chen, S.~Yan, and Jiashi Feng,
\newblock ``Efficient differentiable neural architecture search with meta
  kernels,''
\newblock {\em ArXiv}, vol. abs/1912.04749, 2019.

\bibitem{ref:FewSKD}
Tianhong Li, Jianguo Li, Zhuang Liu, and Changshui Zhang,
\newblock ``Few sample knowledge distillation for efficient network
  compression,''
\newblock {\em 2020 IEEE/CVF Conference on Computer Vision and Pattern
  Recognition (CVPR)}, pp. 14627--14635, 2020.

\bibitem{ref:Cifar}
Alex Krizhevsky,
\newblock ``Learning multiple layers of features from tiny images,''
\newblock 2009.

\bibitem{ref:CifarQuick}
Jasper Snoek, H.~Larochelle, and Ryan~P. Adams,
\newblock ``Practical bayesian optimization of machine learning algorithms,''
\newblock in {\em NIPS}, 2012.

\bibitem{ref:VGG16}
Karen Simonyan and Andrew Zisserman,
\newblock ``Very deep convolutional networks for large-scale image
  recognition,''
\newblock {\em CoRR}, vol. abs/1409.1556, 2015.

\bibitem{Allen19Learning}
Zeyuan Allen-Zhu, Yuanzhi Li, and Yingyu Liang,
\newblock ``Learning and generalization in overparameterized neural networks,
  going beyond two layers,''
\newblock in {\em Advances in Neural Information Processing Systems 32},
  H.~Wallach, H.~Larochelle, A.~Beygelzimer, F.~Alch\'{e}-Buc, E.~Fox, and
  R.~Garnett, Eds., pp. 6158--6169. Curran Associates, Inc., 2019.

\bibitem{neyshabur2018the}
Behnam Neyshabur, Zhiyuan Li, Srinadh Bhojanapalli, Yann LeCun, and Nathan
  Srebro,
\newblock ``The role of over-parametrization in generalization of neural
  networks,''
\newblock in {\em International Conference on Learning Representations}, 2019.

\bibitem{Venturi19Spurious}
Luca Venturi, Afonso~S. Bandeira, and Joan Bruna,
\newblock ``Spurious valleys in one-hidden-layer neural network optimization
  landscapes,''
\newblock {\em Journal of Machine Learning Research}, vol. 20, no. 133, pp.
  1--34, 2019.

\bibitem{Brutzkus2019WhyDL}
Alon Brutzkus and Amir Globerson,
\newblock ``Why do larger models generalize better? a theoretical perspective
  via the xor problem,''
\newblock in {\em ICML}, 2019.

\end{thebibliography}
